\documentclass[pdflatex,sn-vancouver-ay]{sn-jnl}

\usepackage{amsmath,amssymb,amsfonts}%
\usepackage{amsthm}%
\usepackage{mathrsfs}%
\usepackage[title]{appendix}%
\usepackage{xcolor}%
\usepackage{textcomp}%
\usepackage{manyfoot}%
\usepackage{booktabs}%
\usepackage{algorithm}%
\usepackage{algorithmicx}%
\usepackage{algpseudocode}%
\usepackage{listings}%
\usepackage{makecell}
\usepackage{mathtools}
\usepackage[left=3cm,right=3cm,top=2.5cm,bottom=2.5cm]{geometry}

\begin{document}

\title[Article Title]{BuildingGym: An open-source toolbox for AI-based building energy management using reinforcement learning}



\author*[1, 2]{Xilei Dai}\email{xilei.dai@ntu.edu.sg}

\author[1]{Ruotian Chen}
\author[1]{Songze Guan}
\author[1]{Wen-Tai Li}

\author*[1]{Chau Yuen}\email{chau.yuen@ntu.edu.sg}


\affil[1]{School of Electrical and Electronic Engineering, Nanyang Technological University, 50 Nanyang Ave, Singapore, 639798,  Singapore}
\affil[2]{School of Architecture and Urban Planning, Chongqing University, Chongqing, 400045, China}



\abstract{Reinforcement learning (RL) has proven effective for AI-based building energy management. However, there is a lack of flexible framework to implement RL across various control problems in building energy management. To address this gap, we propose BuildingGym, an open-source tool designed as a research-friendly and flexible framework for training RL control strategies for common challenges in building energy management. BuildingGym integrates EnergyPlus as its core simulator, making it suitable for both system-level and room-level control. Additionally, BuildingGym is able to accept external signals as control inputs instead of taking the building as a stand-alone entity. This feature makes BuildingGym applicable for more flexible environments, e.g. smart grid and EVs community. The tool provides several built-in RL algorithms for control strategy training, simplifying the process for building managers to obtain optimal control strategies. Users can achieve this by following a few straightforward steps to configure BuildingGym for optimization control for common problems in the building energy management field. Moreover, AI specialists can easily implement and test state-of-the-art control algorithms within the platform. BuildingGym bridges the gap between building managers and AI specialists by allowing for the easy configuration and replacement of RL algorithms, simulators, and control environments or problems. With BuildingGym, we efficiently set up training tasks for cooling load management, targeting both constant and dynamic cooling load management. The built-in algorithms demonstrated strong performance across both tasks, highlighting the effectiveness of BuildingGym in optimizing cooling strategies.}

\keywords{BuildingGym, Reinforcement learning, Smart building, Demand response control, Flexible building control, Building energy management}



\maketitle

\section{Introduction}
\subsection{Background}

The operations of buildings account for 30\% of global final energy consumption and 26\% of global energy-related emissions according to international energy agency \citep{iea}, indicating a large energy consumption in building operation \citep{hu2022challenges}. Furthermore, building energy consumption is projected to continue rising in the coming years \citep{huo2021dynamic}. Given the significant link between building energy consumption and carbon emissions, optimizing daily operation schedules for building energy systems is essential for achieving global carbon neutrality. A widely recognized strategy for reducing carbon emissions in buildings is to improve the efficiency of HVAC systems, which account for a substantial portion of building energy use \citep{perez2008review}. Building energy simulation serves as an effective tool for enhancing the understanding of HVAC systems and identifying energy-saving potential by optimizing the design and operation of HVAC systems. The following sections present the state-of-the-art in building energy simulation for efficiency, optimal control of building energy systems, and existing toolboxes for building energy management.

\subsection{Building energy simulation for energy saving}

The challenge of optimizing building air conditioning systems exists in several aspects, e.g. the complexity of building physic process \citep{chen2023physics}, the stochastic human occupancy/behavior \citep{yan2015occupant, dai2020review}. Additionally, there are various cooling and heating systems designed for different building types and climate zones. Moreover, new technologies implemented in zero-energy buildings, such as mixed-mode ventilation \citep{peng2022hybrid}, chilled water flow and temperature optimization \citep{trautman2021overall, jia2021optimal}, and thermal and electricity storage \citep{sharma2020economic}, further pose greater challenge for the optimization of building control, necessitating real-time optimal control strategies. To model building physics and simulate energy consumption across various systems, the U.S. Department of Energy developed EnergyPlus \citep{crawley2001energyplus}, an open-source building simulation tool. EnergyPlus has been widely utilized in numerous studies and demonstrates state-of-the-art performance in building energy simulation \citep{boyano2013energy, dahanayake2017studying}.

EnergyPlus has been used in numerous studies for building energy system optimization. \citet{ning2024forced} quantified the energy consumption of a typical office building in Guangzhou equipped with a temperature-humidity-independent control system. \citet{deng2023using} simulate the urban heating energy consumption and quantify the potential energy saving under different policies. \citet{uribe2024development} simulate the energy consumption of buildings with glazing filled with phase change materials. \citet{aruta2023responsive} quantified the energy saving potential of a dynamic envelope. With the help of EnergyPlus, \citet{yan2022optimization} developed an databases for guiding early stages office building design. 
The above research showed that EnergyPlus is an effective tool for building energy simulation. However, for current release of EnergyPlus, it is mainly designed for relatively long-term period simulation and evaluation and is not flexible for real time control optimization. As more and more buildings integrate with the grid, a flexibility control schedule would be helpful for reducing carbon emission from buildings \citep{zhao2025day}. Therefore, it is necessary to develop tools that is able to be used for optimization in control schedule for buildings on the basis of EnergyPlus's ability in simulating building thermal dynamic process. 

\subsection{Building energy system control}

Traditionally, the building energy system is control by rule-based control (RBC), whereby the setpoints for control variables are determined by pre-determined schedules. Even though RBC is easy to implemented in real building energy management system, it scarifies the control performance in energy efficiency because it is not flexible enough for changing factors, such as grid demand signals, outdoor weather etc. Model Predictive Control (MPC) was thus developed that optimizes the control schedule according to the forecast from details physics-based building models. While MPC can be powerful, it falls down due to its requirement in expert knowledge and time for developing reliable physics-based models, as well as the computational complexity to run the optimization for MPC \citep{kaspar2024effects}, which makes the application of MPC in the industry remain limited. To overcome the above-mentioned weakness of RBC and MPC, AI-based solutions have been leveraged to learn the optimal control strategy from building operational data \citep{jia2022deep, lu2023improved}. Compared with traditional control algorithms, reinforcement learning (RL) successfully avoid the cumbersome nature of constructing physics-based models since it is able to learn the optimal control strategy through operation data only. Moreover, as the BMS becoming more and more common in buildings and is able to collect numerous data for RL training, RL becomes a promising solution for smart building control.

Therefore, RL stands out due to its strength in developing autonomous systems with a higher-level understanding of the target environment \citep{arulkumaran2017deep}. RL learns an control agent that determines actions based on the observations through maximizing the reward depending on the objectives. In other words, RL learns the optimal agent through trial and errors. Due to this strength, RL does not require system-wide optimization at every time step to make optimal decisions \citep{mocanu2018line}. Previous studies have shown outstanding performance of RL in building energy system optimization. For instance, \citet{dai2023deciphering} developed RL control algorithms for mixed-mode ventilation, leveraging building thermal dynamics to significantly reduce cooling energy consumption by maximizing the duration of natural ventilation while maintaining a satisfactory indoor thermal environment. Similarly, \citet{chen2018optimal} optimized window status schedules using RL, achieving energy savings of 13\% and 23\% in mixed-mode ventilation buildings located in Miami and Los Angeles, respectively. \citet{shen2022multi} created a multi-agent control strategy employing a double deep Q-network for building energy systems with renewable energy. When applied to an office building, this strategy reduced energy costs by 8\% compared to conventional rule-based control methods. Additionally, \citet{deng2022towards} utilized deep Q-networks (DQN) in a multi-zone control task, achieving improvements of 13\% in energy savings and 9\% in thermal comfort. Moreover, they also reported that the proposed method shows good generalization to an unseen building environment. Regarding demand response control, \citet{xie2023multi} implemented multi-agent attention-based deep RL, effectively reducing net load demand. Through advanced transfer learning technologies, RL control strategy can be successfully implemented in real buildings \citep{Hilo}. According to these researches, it can be seen that RL performs well in building control with different control targets. However, finding a unified control strategy applicable to all buildings remains challenging due to the complexity of building physics and the diversity of control targets. It is necessary to simplified the developing process of RL strategy for building energy system control. 

\subsection{Existing toolbox/environment for building energy management}

To facilitate the training of the RL algorithms, a viable solution is to develop an easy-to-use framework featuring well-tested AI algorithms, enabling researchers to focus on optimizing control strategies rather than spending time on the implementation of AI techniques. Several studies have begun developing toolboxes and packages to facilitate RL training. For instance, \citet{vazquez2020citylearn} developed CityLearn to provide a standardized training environment for testing and comparing different types of RL algorithms. However, CityLearn relies solely on simulation results from EnergyPlus to train deep learning models as surrogate models for building physics processes, without directly integrating EnergyPlus as the simulator during training. Similarly, \citet{wang2021alphabuilding} proposed AlphaBuilding ResCommunity, which aims to offer a standardized multi-agent virtual testbed for community-level load coordination. They also utilize reduced-order models instead of EnergyPlus to simulate the thermodynamics of thermostatically controlled loads. While using a reduced-order model or surrogate deep learning strategy can accelerate the training process and is suitable for building-side demand response at the community level, it compromises the ability to control equipment and systems effectively. Additionally, it necessitates a dedicated deep learning model for building physics. Zhang developed Gym-Eplus for RL training within EnergyPlus \citep{epp-gym}. However, Gym-Eplus does not provide an application programming interface (API) that connects with environments outside of EnergyPlus, meaning the Gym-Eplus environment operates as a standalone EnergyPlus model. These limitations hinder its applicability in community demand response control, where buildings must respond to external signals, such as in Building-EV communities \citep{liao2024comparative, sridharan2023hybrid}. To facilitate the setup of simulation environment, \citet{blonsky2021ochre} proposed an object-oriented and controllable residential energy model, OCHRE, built on RC model as the energy simulator. However, the RC-based modeling framework may reduce the accuracy of energy model and make some system-side control unavailable. One possible solution is to develop an toolbox that is convenient to conduct co-simulation between detailed physic-based simulator, e.g. EnergyPlus, with RL training environment. To achieve this, we developed BuildingGym as a toolbox for AI-driven building energy management, offering a user-friendly interface to interact with the EnergyPlus API and set up customized RL training tasks efficiently. Table \ref{tab:features} compares the features between the proposed BuildingGym and some existing tool.

\begin{table}[!h]
    \centering
        \caption{The comparison of the features between BuildingGym and existing tools for building energy management/optimization}
    \label{tab:features}
    \begin{tabular}{lccccc}
    \toprule
    
         & \makecell{Physic-based\\simulator} & \makecell{Customized \\ building model}  &\makecell{External \\observation}  & \makecell{Built-in\\ RL} & \makecell{Customized\\ RL} \\
    \midrule
       CityLearn\citep{vazquez2020citylearn}  & - & - & - & $\bullet$ & $\bullet$ \\
       Gym-Eplus\citep{epp-gym}  & Detailed & $\bullet$ &-  & $\bullet$ & - \\
       OCHRE\citep{blonsky2021ochre}  & Reduce-order & - & - & - & - \\
       AlphaBuilding\citep{wang2021alphabuilding}  & Reduce-order & - & - & $\bullet$ & $\bullet$ \\
       \textbf{BuildingGym}  & Detailed & $\bullet$ & $\bullet$ & $\bullet$  & $\bullet$ \\
    \bottomrule
    \end{tabular}

\end{table}

\subsection{Literature gap and research objectives}

Control strategies for HVAC systems in building can be optimized to reduce carbon emission from building sectors. We noted the growing research related to RL in building energy system control, and we identify three key gaps in the literature as: 1) difficulty of using customized building environments as environment for RL training; 2) easy-to-use toolbox to test RL algorithms for common building energy control problem; 3) community-level environment that is able to accept external control signal from up-level energy systems for building energy system control.

To address the aforementioned research gap, we propose BuildingGym for AI-based building energy management. The aim of BuildingGym is thus to address these key gaps, and as such the primary objectives are to:

\begin{itemize}
    \item Extend the physics-based simulator to the general environment for RL training that can be used for a wide range of building energy system optimization tasks;

    \item Close the gap between building managers and AI specialists by providing standardized RL training toolbox for building energy management;
    
    \item Enable building energy systems to receive external control signals from communities, enhancing integration with future grid-responsive building through BuildingGym;
    
    \item To demonstrate how BuildingGym can be used to develop and test for building control problem, and validate the effectiveness of BuildingGym for demand response control optimization.

\end{itemize}

The remaining sections of this paper are organized as follows: Section 2 presents the architecture and key modules of BuildingGym. Section 3 evaluates the performance of the proposed RL-based control optimization toolbox in managing cooling loads. Section 4 discusses the features and significance of BuildingGym for AI-based building energy management, as well as potential directions for future research. Finally, conclusions are drawn in Section 5.

\section{Proposed BuildingGym}

This section first introduces the overall architecture design of the BuildingGym. Generally speaking, users only need to complete the basic configuration in the User Configuration Module to implement the RL training for user-specific tasks. After completing the basic configuration, BuildingGym calls the training modules using the corresponding algorithm module. The training process can be visualized using Wandb package \citep{wandb} and the good control agents for the user-specific task will be saved automatically if the criteria is defined. To make the overall workflow clear, we also provide a virtual testbed and corresponding demo examples to show the performance of built-in RL for cooling load management control problems in the smart building fields. The code of BuildingGym and its application examples is publicly available in GitHub \citep{buildinggym}, as well as the detailed documentation for user participation and contribution. (note: we will provide the link in the paper and make the repository publicly available once the paper is accepted.)

\subsection{BuildingGym Architecture}

\begin{figure}
    \centering
    \includegraphics[width=.95\linewidth]{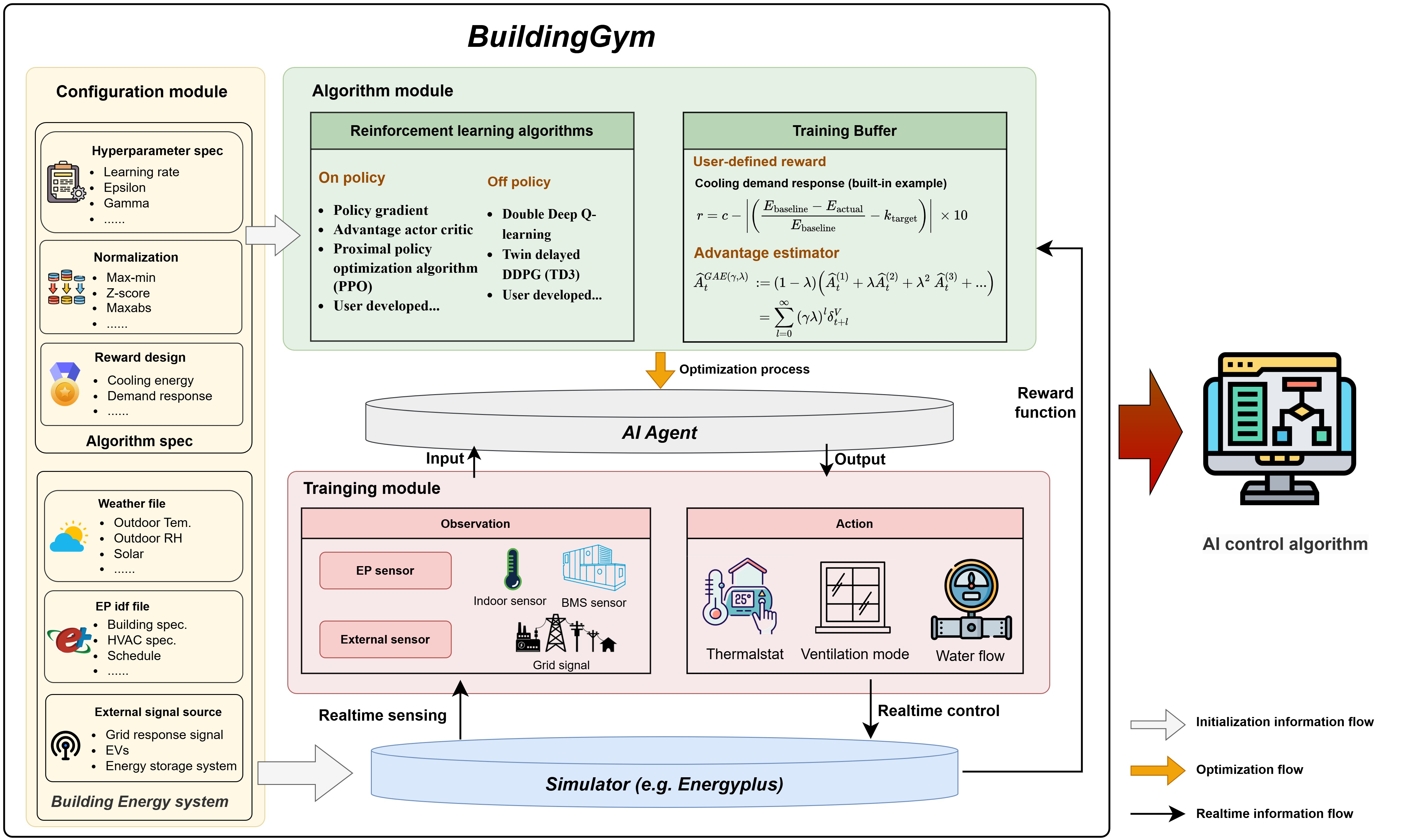}
    \caption{The architecture of BuildingGym, which includes configuration module, training module and algorithm module}
    \label{BuildingGym_archi}
\end{figure}

To realize the simple configuration for RL training in building energy simulator, we propose BuildingGym that consists of three main modules: the configuration module, the training module, and the algorithm module. The overall architecture of BuildingGym is illustrated in Fig. \ref{BuildingGym_archi}. Generally speaking, The configuration module provides unified interfaces for users to configure the building energy model, observation/action space, and hyper-parameters required for RL training. The training module runs the building energy simulation for user-specific model, conducting observation and action through EnergyPlus API. Meanwhile, the RL control agent is trained by calling the algorithm module during the simulation. The algorithm module defines the training workflow, including both on-policy and off-policy RL. The detailed description for these modules is as follows:

\begin{itemize}
    \item  The configuration module serves as a template to guide users through the setup process. A unified configuration interface is defined which makes sure the simplicity to test different RL algorithms for building models. Once users complete the configuration, they can proceed to training their own RL for the specific building.

    \item  The training module integrates the EnergyPlus engine and the RL algorithm module to initiate training. Users can visualize the training process online based on their configurations. The best-performing control agents, which achieve the highest rewards, will be automatically saved for future use. A unified output structure is defined to make sure all training results are always stored in a consistent form and thus can be easily fed to various post-analysis tools.

    \item  The algorithm module defines the training workflow, encompassing both on-policy and off-policy training. Users have the option to develop their own RL algorithms or utilize the built-in common RL algorithms. 
\end{itemize}

Comparing to previous platforms, BuildingGym provides user-friendly interface for configuration on RL training for self-defined problems. Moreover, BuildingGym can take external signals into consideration for control, instead of only using observation from building energy simulators. This feature makes BuildingGym applicable to grid-interactive buildings under future complex energy systems where buildings may response to external signals. Last but not least, the simulation engine in BuildingGym is replaceable to other simulation engines or connecting to actual building BMS system, providing more possibilities for implementation in real buildings.

\subsection{Configuration module}

The configuration module is designed to help users complete the necessary setup for training the desired RL agent tailored to their specific tasks. Table \ref{configuration_summary} summarizes the information required from users, which is organized into three main components: building the energy model, formulating the control problem, and specifying the algorithm.

\begin{table}[htbp]
\caption{The summary of configuration information from user in BuildingGym}\label{configuration_summary}
\begin{tabular}{lll}
\toprule
                      & Configuration information & Type                                                            \\ \midrule
Problem formulation   & Internal observation      & EnergyPlus sensor                                               \\
                      & External observation*     & Self-developed function for real time external signal             \\
                      & Reward function           & Class function in environment class of BuildingGym              \\
                      & Action space              & EnergyPlus actuator                                             \\
                      & Action function*          & Class function in environment class of BuildingGym              \\
Building simulator    & Building model            & idf file                                                        \\
                      & Weather file              & epw file                                                        \\
Algorithm specification & RL algorithm              & BuildingGym object, either built-in or self-developed algorithm \\
                      & Training hyper-parameter   & DataClass object, with default value                            \\ 
                      & Normalization*            & Normalization method for input observation                  \\   \bottomrule
\end{tabular}
\begin{tablenotes}
    \footnotesize
    \item [a] *Optional information
\end{tablenotes}
\end{table}

In the building energy model section, users need to provide the EnergyPlus model along with the corresponding weather file to establish the complete environment for RL. By default, a standard office building energy model, provided by the U.S. Department of Energy, is used as the built-in virtual testbed \citep{model-1, model-2}.

In the control problem formulation, the key components are the observation space, action space, action function, and reward function. BuildingGym allows RL algorithms to get observations from both EnergyPlus sensors and external sensors. This capability enables communication between the building energy system and external entities, such as grid requirements and EV charging status, making BuildingGym suitable for a wider range of control problems beyond treating buildings as standalone entities. For the action space, BuildingGym supports control of all actuators in EnergyPlus, enhancing its applicability to various building energy system control challenges. Figure S1 provides the example of look-up table for available actuator and sensor in EnergyPlus provided by BuildingGym.

The action function and reward function can be user-defined. By specifying the action function, users can determine how to translate the RL output into actuator commands at each time step. By default, the RL output is directly used as the actuator command. The reward function is crucial for guiding RL training. We provide a demonstration reward function in our demo case for the demand response control problem.

As we know, fine-tuning RL algorithms is crucial for maximizing their control capabilities. In the algorithm specification, we consolidate all hyperparameters into a single template file for each built-in RL algorithm, along with default settings for the demo building energy management problem. This setup allows users to easily modify all hyperparameters in one file, streamlining the fine-tuning process. As a result, users can more effectively leverage the RL algorithm to develop control agents with satisfactory performance.

\begin{figure}
    \centering
    \includegraphics[width=.7\linewidth]{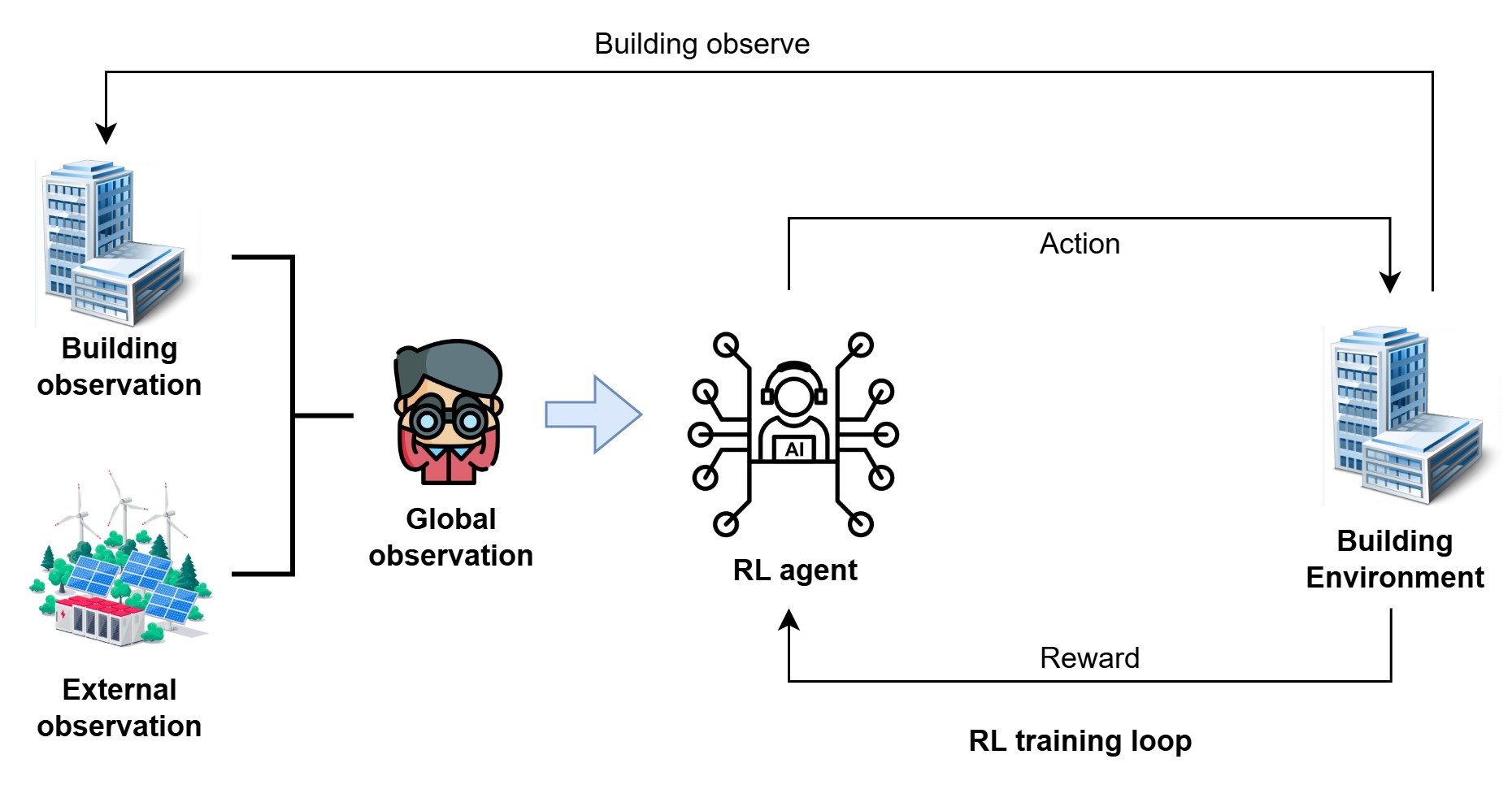}
    \caption{The implementation overview of receiving external variables functions. The building observation is from simulator and the external observation can be defined by users to simulate any external signals.}
    \label{fig:ext_obs}
\end{figure}

Figure \ref{fig:ext_obs} shows the implementation overview of receiving external variables functions. By enabling BuildingGym to be able to accept external variables as input features, it can make BuildingGym applicable to grid-interactive buildings under future complex energy systems, which is important for future smart buildings. To achieve this, we designed a scheme to make RL able to accept external variables from communities instead of only considering the observations provided by building energy simulators. To enable this function, users only need to activate the external variable interface by setting $ext\_obs\_bool$ as True in BuildingGym and then provide the self-define functions to provide external variable at each time step. 

\subsection{Training module}

The training module serves as the crucial link between the configuration and algorithm modules. In this module, users specify the type of RL algorithm they wish to use. The training module retrieves the corresponding settings from the configuration module and establishes the training environment for RL. During the training process, operational data (observation and action at each time step) is collected in a data buffer through callback function, and the specified RL algorithm is employed to train the agent.

\subsubsection{Interaction with environment}

To setup the training environment in Python, we utilized EnergyPlus State API \citep{state-api} to run the EnergyPlus for building energy simulation in Python environment, which enables Python environment to create and manage state instances for using EnergyPlus API methods. When the callback function is defined for observation and action at each time step, the current state instance is passed as the only argument, which allows Python code to close the loop and pass the current state when making API calls inside callbacks.

The callback function in EnergyPlus is utilized to make sure the observation data can be collected and the RL agent can make decision according to the observation and pass to EnergyPlus to simulate the control. EnergyPlus provides 21 callback functions in Runtime API \citep{callback} for interacting with simulation engine at different time points in simulation. In BuildingGym, the default callback point is at the end of zone time step and after zone reporting. Users can change the callback points by selecting the suitable callback functions according to the specific control tasks. 

It should be noted that the EnergyPlus can be replaced by any building energy simulator if it can provide API for observation and action at each time step.

\subsubsection{Output of training module}

The output of the training module is the control agent and the corresponding simulation results underlying the same agent. The control agent is formatted as an neural network in PyTorch format \citep{torch}. By using the standard format, user can easily load the trained agent for other purposes. Also, the BuildingGym provide function to load the pre-trained agent for transferring training or evaluation.

\begin{figure}[hb]
    \centering
    \includegraphics[width=0.9\linewidth]{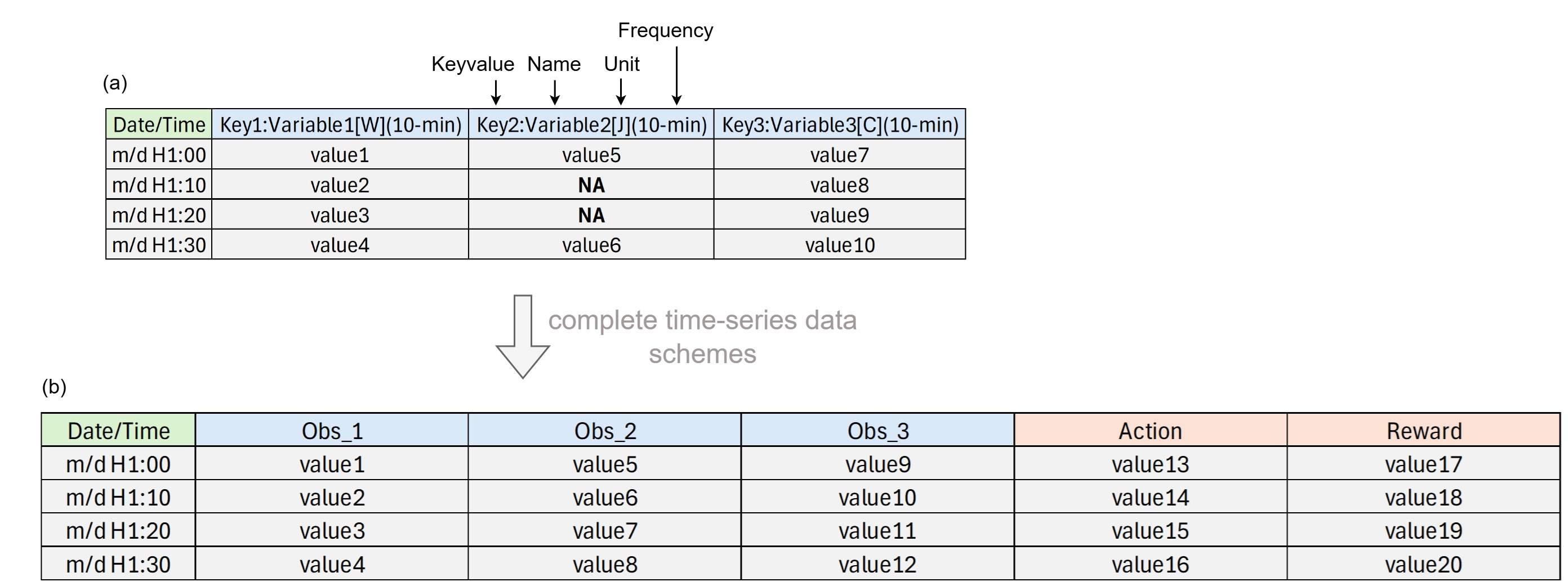}
    \caption{An example of output CSV table format in BuildingGym where Table (a) is the standard CSV table format from EnergyPlus and Table (b) is the output data with complete time-series of observation, action and reward}
    \label{fig:data}
\end{figure}

The simulation results are formatted as CSV table following the complete time-series data schemes. The complete time-series data schemes in BuildingGym is designed to extract and represent EnergyPlus simulation result into complete tables, aiming to provide a standardized way of organizing data that makes it easier to manipulate, analyze, and visualize. The key idea is that data should be structured consistently and completely so that tools for data analysis can work efficiently.

Table (a) in Figure \ref{fig:data} shows an example format from EnergyPlus CSV output table, while Table (b) shows the corresponding complete time-series representation of the same simulation period format following the complete time-series data schemes. Although the format of Table (a) is storage-efficient by fully crossed designs, it violates with the complete time-series principles that include missing values (NA in row 2 and 3) in the table. In practice of BuildingGym, the observation and action data is recorded at every callback points following the complete time-series data scheme. Moreover,  since the reward data is an important evaluation metric for reinforcement learning, it is also recorded in the Table (b). Taking together, Table (b) forms a complete time-series data table. Even such data format may require larger memory, the advantage of saving time in cleaning simulation results makes the complete time-series data scheme that structures data in a standard and straightforward way stands out.

\subsection{Algorithm module}

The algorithm module is the core component responsible for implementing RL algorithms in BuildingGym. It supports both on-policy and off-policy training strategies. The on-policy training strategy includes policy gradient methods, advantage actor-critic, and proximal policy optimization algorithms. Off-policy training encompasses deep Q-learning and twin delayed DDPG algorithms. BuildingGym allows users to either utilize built-in algorithms or implement customized algorithms within the two training strategies. The built-in RL algorithms are developed with reference to several open-source RL libraries, including Stable Baselines3 \citep{raffin2021stable}, CleanRL \citep{huang2022cleanrl}, and Gymnasium \citep{towers_gymnasium_2023}.

\begin{figure}
    \centering
    \includegraphics[width=0.9\linewidth]{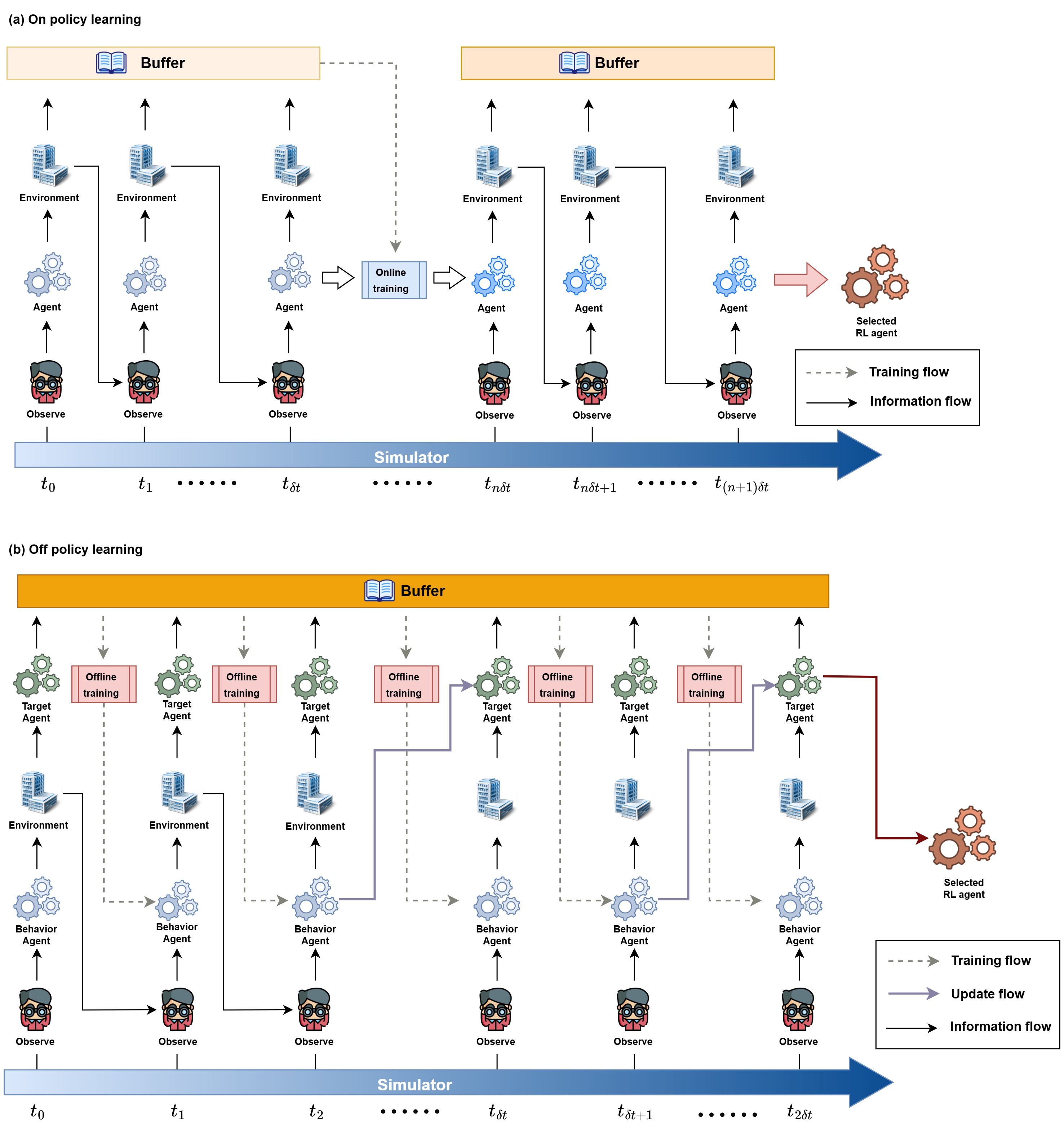}
    \caption{The schematic diagram of on-policy learning training workflow}
    \label{on_policy}
\end{figure}

In on-policy learning, the agent learns and refines the policy it is currently following. During training, the agent interacts with the environment using this policy, collects data, and updates the same policy based on the gathered information. This characteristic often results in a faster convergence speed compared to off-policy algorithms. In off-policy learning, the agent learns a policy that differs from the one it is currently following. The policy being optimized (the target policy) is different from the policy used to generate the data (the behavior policy). During the training process, the agent can use data collected from any policy (or even from an experience replay buffer) to update its policy. This characteristic enables the agent to learn from past experiences or from the experiences of other agents. The training workflow for on-policy and off-policy learning in BuildingGym is illustrated in Fig. \ref{on_policy}. In Section \ref{sec:rl_math}, we will present the preliminary knowledge of the RL algorithms and in Section \ref{sec:rl_example}, we will present how to complete the configuration in BuildingGym for a specific control task.




\subsection{Demo case study in BuildingGym}

To validate the effectiveness of BuildingGym, we conducted two demo studies in two virtual experiments. These experiments are designed to evaluate the performance of the built-in RL algorithms across two common control tasks. The first control task aims to maintain or decrease cooling energy within a constant desired range, enabling buildings to respond more precisely to grid demand. The second control task is similar to the first one, but the desired reduction range varies over time based on external grid signals. This task is expected to become increasingly common and important in the near future due to the growing demand for real-time demand response as more renewable energy sources are integrated into the energy system. Testing this control task demonstrates BuildingGym's capability to integrate both internal and external variables for effective building energy management.

\begin{figure}
    \centering
    \includegraphics[width=0.5\linewidth]{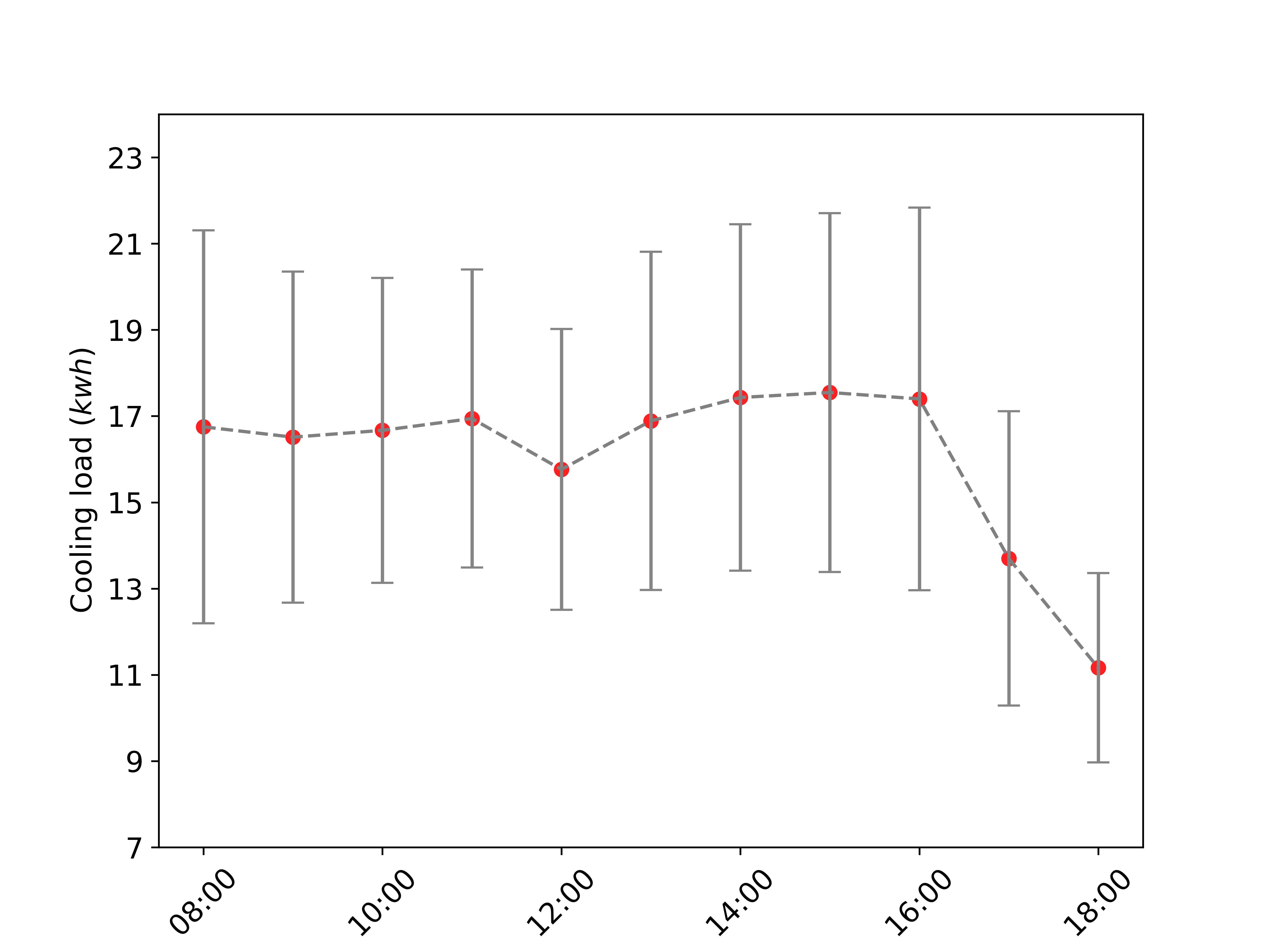}
    \caption{The hourly average cooling energy and the standard deviation during working time}
    \label{fig:baseline}
\end{figure}

The building tested in the demo cases is the standard small office building energy model from the dataset for commercial buildings based on CBECS data \citep{epmodel}. The specifications of the building are detailed in Table \ref{spec}. The simulation period runs from August 1st to 31st, considering the weather conditions in Miami. The corresponding cooling energy under the baseline cooling setpoint schedule for August is shown in Figure \ref{fig:baseline}. The hourly cooling energy from 8:00 AM to 4:00 PM averages about 17 kW, with a standard deviation of approximately 3 kW throughout most of the day, indicating significant variation across different days. Given that such uncertainty may pose challenges for grid load management, it is important to explore a more accurate cooling load management strategy by regulating the cooling setpoints.
\begin{table}[!h]
\centering
\footnotesize
\caption{The specification of built-in virtual testbed}\label{spec}
\begin{tabular}{p{6cm}p{6cm}}
\toprule
Feature &  Attribution\\
\midrule
No. of floors&1\\
Total floor area $(m^2)$&511.16\\
Ceiling Height $(m)$&3.05\\
Window-Wall Ratio (\%)&15.89\\
Roof construction&Typical iead roof r-13.51 adj roof insulation adj roof insulation\\
Wall constructions&Typical interior wall\\
Exterior walls&Typical insulated exterior mass wall adj ext wall insulation adj ext wall insulation\\
Air terminals&Single duct, constant volume\\
Occupancy ($m^2$/person)&23\\
Lighting ($W/m^2$)&15.59\\
Electricity Equipment ($W/m^2$)&8.07\\

\bottomrule
\end{tabular}
\end{table}

\section{Preliminary knowledge of deep reinforcement learning}\label{sec:rl_math}

In this section, we briefly describe the background of deep reinforcement learning for each algorithm, which builds up the foundation for the proposed BuildingGym for building energy management.

In a typical RL, the agent gradually predicts its best action through the trial-and-error interactions with the environment over time, applying actions to the environment, observing the instant rewards and the transitions of state of the environment. The RL learning process can be characterized by several elements, including the observation, the action, the instant reward, the policy and the value function:

\begin{itemize}
\item\textbf{Observation:} A partial or complete representation of the environment's state that the agent uses to make decisions, denoting $s_t \in S$ as observation at each time step. 

\item\textbf{Action:} A decision that the agent takes in response to its current observation of the environment.  Once the agent takes an action $a_t \in A$ at time $t$ following a policy $\pi$, the agent gets a reward $r_t$

\item\textbf{Reward:} A reward $r_t$ provides feedback to the agent about the quality of its actions. It serves as the primary learning signal, guiding the agent toward desirable behaviors by encouraging actions that maximize cumulative rewards over time.

\item\textbf{Policy function:} Defines the agent's strategy for selecting actions based on observations from the environment, denoting as $\pi(s)$.

\item\textbf{Value function:} Predict expected cumulative reward an agent can obtain from a given state or state-action pair, denoting as $V(s,a)$.

\item\textbf{Experience:} Defined as $\{s_t, a_t, r_{t}, s_{t+1}\}$. The experience are used to optimization the policy/value function in RL.
\end{itemize}

In BuildingGym, we provide friendly interface for defining the observation, the action and the reward function. The policy and value function will be created automatically accordingly.

For RL algorithms, the fundamental objective is to optimize reward accumulation by employing a specific strategy or policy. In the following sections, we present the key mathematical foundations underlying each RL algorithm implemented in BuildingGym.

\textbf{Policy gradient}: The policy gradient algorithm aims to model and optimize the policy directly. It typically takes the observation state as input and outputs the proposed action. The policy is often represented by a neural network parameterized by \(\theta\), denoted as \(\pi_\theta(a|s)\). The value of the reward function depends on this policy and then common optimization algorithms for neural network training can be applied to optimize for the best expected reward. The expected reward can be written in Eq \ref{exp_r}.

\begin{equation}
    J(\theta) = \sum_{s \in S}d^\pi(s)\sum_{a\in A} \pi_\theta(a|s) Q^\pi(s,a) \label{exp_r}
\end{equation}
and its gradient expression under Monte-Carlo sampling can be written as:
\begin{equation}
    \nabla J(\theta) \approx \frac{1}{N} \sum_{\tau \in N} \sum_{t=0}^{T-1}\nabla_\theta log\pi_\theta(a_t, s_t)R(\tau)
\end{equation}

\textbf{Advantage actor critic}: As indicated by its name,  this algorithm consists of two main components: the actor and the critic.  The actor functions similarly to that in policy gradient methods, taking the state as input and directly outputting the corresponding actions. However, the key difference is that the gradient in A2C is not estimated using discounted rewards. Instead, it employs the generalized advantage estimator \(GAE(\gamma, \lambda)\) to calculate the gradient of the actor during training. The advantage is computed as the exponentially-weighted average of these k-step estimators, as shown in Eq. \ref{adv}\citep{schulman2015high}.

\begin{align}
           \hat{A}_t^{GAE(\gamma, \lambda)}  &\coloneq (1-\lambda)\left( \hat{A}_t^{(1)} + \lambda \hat{A}_t^{(2)} + \lambda^2  \hat{A}_t^{(3)} + ...\right) \\
           &=(1-\lambda) \left( \delta_t^V\left(\frac{1}{1-\lambda}\right) +
           \lambda\delta_{t+1}^V\left(\frac{\lambda}{1-\lambda}\right) +
           \lambda^2\delta_{t+2}^V\left(\frac{\lambda^2}{1-\lambda}\right) + ...\right)\\
           &= \sum_{l=0}^\infty(\gamma\lambda)^l\delta_{t+l}^V
           \label{adv} 
\end{align}
where $A_t^k$ is the sum of $k$ of these TD residual of $V$ with discount $\gamma$:

\begin{align}
    \hat{A}_t^{(1)} &\coloneq \delta_t^V = -V(s_t) + r_t + \gamma V(s_t+1)\\
    \hat{A}_t^{(2)} &\coloneq \delta_t^V + \gamma \delta^V_{t+1} = -V(s_t) + r_t+ \gamma r_{t+1} + \gamma^2 V(s_t+2)\\
    \hat{A}_t^{(3)} &\coloneq \delta_t^V + \gamma \delta^V_{t+1} + \gamma^2 \delta^V_{t+2} = -V(s_t) + r_t+ \gamma r_{t+1} + \gamma^2 r_{t+2} + \gamma^3 V(s_t+3)\\
    \hat{A}_t^{(k)} &\coloneq \sum_{l=0}^{k-1}(\gamma\lambda)^l\delta_{t+l}^V = -V(s_t) + r_t + \gamma r_{t+1} + ... +  \gamma^{k-1}r_{r+k-1} + \gamma^kV(s_t+k)
\end{align}
So the actor loss and the critic loss are calculated as Eq. \ref{a2c_loss}:
    \begin{align}
        Loss_{actor} = -GAE(\gamma, \lambda) * log \pi(a,s) \label{a2c_loss}\\
        Loss_{critics} = \frac{1}{n}\sum_{i=1}^n(R(\tau, i) - V(s_t, i))^2
    \end{align}

\textbf{Proximal policy optimization algorithm (PPO)}: This RL algorithm is a actor-based on-policy algorithm. The goal of PPO is to take the largest possible improvement step  to the policy using the currently available data. The PPO update the actor, $\pi_{\theta}$,via Eq \ref{PPO_loss}:

\begin{equation}
    \theta_{k+1} = \text{arg}\max_\theta \mathbb{E}_{s,a\sim \pi_{\theta_k}} \left[ L\left(s,a,\theta_k,\theta)\right)\right] \label{PPO_loss}
\end{equation}
where
\begin{equation}
    L\left(s,a,\theta_k,\theta)\right) = min\left( \frac{\pi_\theta(a|s)}{\pi_{\theta_k}(a|s)}A^{\pi_{\theta_k}(s,a)}, \text{clip}\left( \frac{\pi_\theta(a|s)}{\pi_{\theta_k}(a|s)}, 1-\epsilon, 1+\epsilon \right)A^{\pi_{\theta_k}(s,a)} \right)
\end{equation}



\textbf{Double Deep Q-Learning}: This RL algorithm is a critic-based method. It does not have an actor that directly outputs the desired actions. Instead, the algorithm makes decisions based on the critics, which estimate the expected accumulated rewards, \(Q(s, a)\), given the state and actions. In other words, the Q-learning algorithm aims to predict the Q value for each action under a given state, selecting the action that is expected to yield the highest Q value as the predicted action. During training, the Double Deep Q-Learning algorithm seeks to train a neural network that can accurately predict \(Q(s, a)\), which is defined in Equation \ref{q_value}.
\begin{equation}
    Q^\pi(s,a) = r + \gamma Q^\pi(s', \pi(s')) \label{q_value}
\end{equation}
The difference between the two sides of the equality is known as the temporal difference error, \(\delta\). During training, the temporal difference error is treated as the loss, which is intended to be minimized by the optimization algorithm.
\begin{equation}
    \delta = Q(s,a) - (r + \gamma \max_a Q(s',a))
\end{equation}

\textbf{Twin Delayed DDPG (TD3)}: This RL algorithm includes both an actor, \(\mu_\theta\), and critics, \(\phi_n\), in the agent. The training of the critics is similar to Q-learning, which formalizes the learning target using the Q-function (Eq. \ref{q_value}). However, TD3 incorporates an actor with clipped noise added to predict target actions, as shown in Equation \ref{td3_actor}. This smoothing operation for the actor can be considered a regularizer for the algorithm. By doing so, the actor can offset the bias introduced by the Q-function in certain situations. For instance, a common failure mode in deep Q-learning algorithms is when the Q-function predicts an incorrect sharp peak for some actions. In such cases, the actor is able to quickly exploit that peak value.
\begin{equation}
    a'(s') = \text{clip}(\mu_{\theta_\text{target}}(s') + \text{clip}\left(\epsilon, -c, c), a_{\text{low}}, a_{\text{high}}\right), \hspace{6pt} \epsilon \sim \mathcal{N}(0,\sigma) \label{td3_actor}
\end{equation}

For training the critic, a trick of TD3 is to learn multiple Q-functions instead of one, and use the smallest Q-values to formulate the loss functions:
\begin{equation}
    y(r, s', d) = r + \gamma (1-d) \min_{n_{\text{critic}}} Q_{\phi_n}(s', a'(s'))
\end{equation}
and then the critics $\phi_n$ are learned by regressing to target in Eq. \ref{td3_Q_target} and the policy is learned by maxing $Q_{\phi_n}$ as shown in Eq. \ref{td_3_policy}.
\begin{align}
    &L(\phi_n, \mathcal{D}) = \mathbb{E}_{(s,a,r,s',d)\sim\mathcal{D}}\left[ \left( Q_{\phi_n}(s,a) - y(r,s',d) \right)^2 \right]\label{td3_Q_target}\\
    &\max_\theta \mathbb{E}_{s\sim \mathcal{D}}\left[ Q_{\phi_1}(s, \mu_\theta(s)) \right]\label{td_3_policy}
\end{align}

\section{Applications of BuildingGym for cooling load management}

In this section, we use BuildingGym to train and test the performance of built-in RL algorithms on constant (Section \ref{constant_demand}) and dynamic (Section \ref{dynamic_demand}) demand reduction for cooling loads. The configuration of each control problem, training process, and overall performance are presented. The training process of all RL algorithms is shown in Figure S2, and the corresponding fine-tune parameters for training are listed in Table S1 in the supplementary material.


\subsection{Cooling load management with constant target}\label{constant_demand}

With the share of renewable energy increasing, buildings may be requested to reduce their electricity usage to some extent sometimes since the production of renewable energy is not as stable as traditional energy sources, especially during extreme weather events. For this demand response task, achieving an accurate reduction in cooling load is crucial since both excessive and insufficient reductions can negatively impact the grid during the demand response period. In another aspect, HVAC systems typically consume a significant amount of energy in buildings, and their energy usage can be effectively reduced by adjusting indoor cooling setpoints. This makes HVAC systems a potential control target for responding to grid demand requirements.

In this control task, we evaluate the performance of BuildingGym in managing cooling energy reduction. The control variable is the indoor cooling setpoint, and the objective is to reduce cooling energy as close to the desired level as possible—neither too much nor too little. Failure to achieve this balance could compromise the demand management requirements set by the grid. Furthermore, excessive reductions in cooling energy may lead to significant losses in indoor thermal comfort due to elevated temperatures, decreasing occupants' willingness to participate in demand management.

\subsubsection{Basic configuration}\label{sec:rl_example}

\begin{figure}[htb]
    \centering
    \includegraphics[width=.9\linewidth]{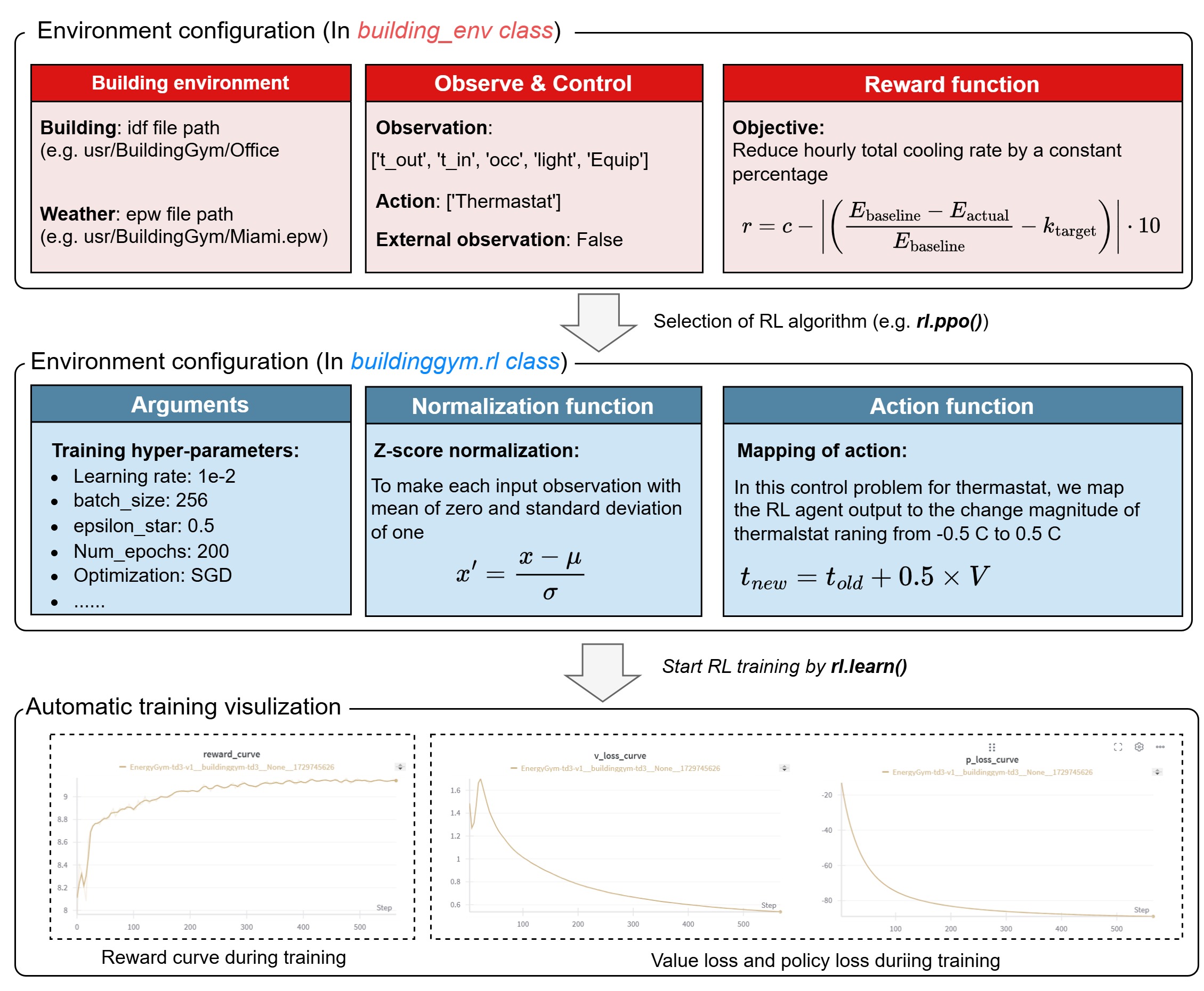}
    \caption{The overview of configuring a training task in BuildingGym. The \textit{\textbf{building\_env}} class serves as the interface for environment configuration and problem formulation, the \textit{\textbf{buildinggym.rl}} class serves as the interface for RL algorithm configuration}
    \label{fig:training}
\end{figure}

Figure \ref{fig:training} shows the overall workflow of BuildingGym to configure a RL training task for cooling load management. BuildingGym provides friendly interface for user to complete environment configurations, as shown by the example of key code in Appendix. The entire source code for the configuration and training is available at our GitHub page.

For this control task, the objective is to reduce hourly total cooling rate,($E_{actual}$) by a constant percentage ($k_{target}$) comparing to the baseline total cooling rate ($E_{baseline}$). The reward function is thus defined as follows:
\begin{equation}
    r = c - \left|\left(\frac{E_{\text{baseline}} - E_{\text{actual}}}{E_{\text{baseline}}} - k_{\text{target}}\right)\right|*10
\end{equation}
where $c$ is the constant value added to the reward to make the RL easier to converge, $k_{target}$ is the target reduction percentage compared to the baseline energy consumption.

The input observations for all RL algorithms are outdoor air temperature, indoor air temperature, and occupancy/lighting/equipment schedules, which are key parameters that significantly impact building cooling load. In regards to the action function, the output action space is defined as a discrete set \{-1, 0, 1\}, where "0" means no change to the cooling setpoint, "-1" represents a decrease of 0.5 $^\circ$C, and "1" signifies an increase of 0.5  $^\circ$C for policy gradient, A2C, PPO, and DQN. For the TD3 algorithm, the action output is a continuous value ranging from -0.5 $^\circ$C to 0.5 $^\circ$C, representing the change in cooling setpoint for the next time step. To facilitate the training and keep cooling setpoint in acceptable range, we manually set a constrain to make sure the cooling setpoint between 23 $^\circ$ and 27 $^\circ$. The timestep for observation and control is set as 10 min.

\subsubsection{Control accuracy for constant target}

In this section, we present the performance of trained RL agent for cooling load management with constant target to validate the effectiveness of BuildingGym in configuring RL training for common building energy system optimization.

\begin{figure}[h]
    \centering
    \includegraphics[width=0.8\linewidth]{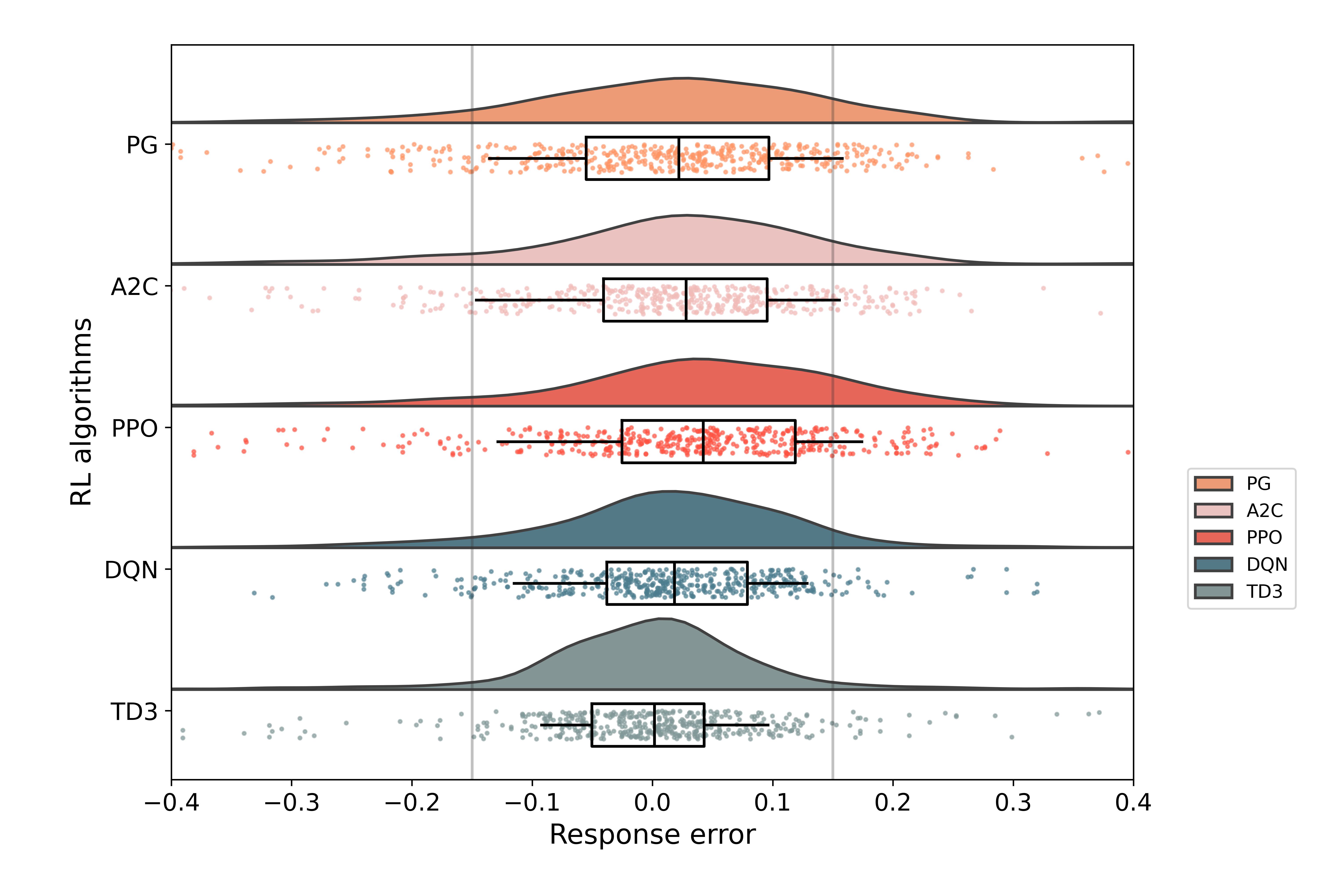}
    \caption{Overall performance of each RL algorithms for constant demand response control}
    \label{constant_boxplot}
\end{figure}

Figure \ref{constant_boxplot} shows the overall performance of demand response control, which is defined as the difference between actual reduction and target reduction in percentage. The median response error is within 1\% for PG, A2C, and TD3, and is about 1.5\% for DQN and PPO. This indicates that all RL algorithms show good performance in demand response control in terms of long-term demand response control. In another aspect, the two off-policy RL algorithms are also able to make the 10th and 90th percentile within $\pm$15\%, while the 90th percentile of three on-policy RL algorithms is slightly higher than 15\%. Among all RL algorithms, TD3 offers the best performance, with a median control error of 0.5\% and the smallest range between the 10th and 90th percentiles.

\begin{figure}
    \centering
    \includegraphics[width=0.6\linewidth]{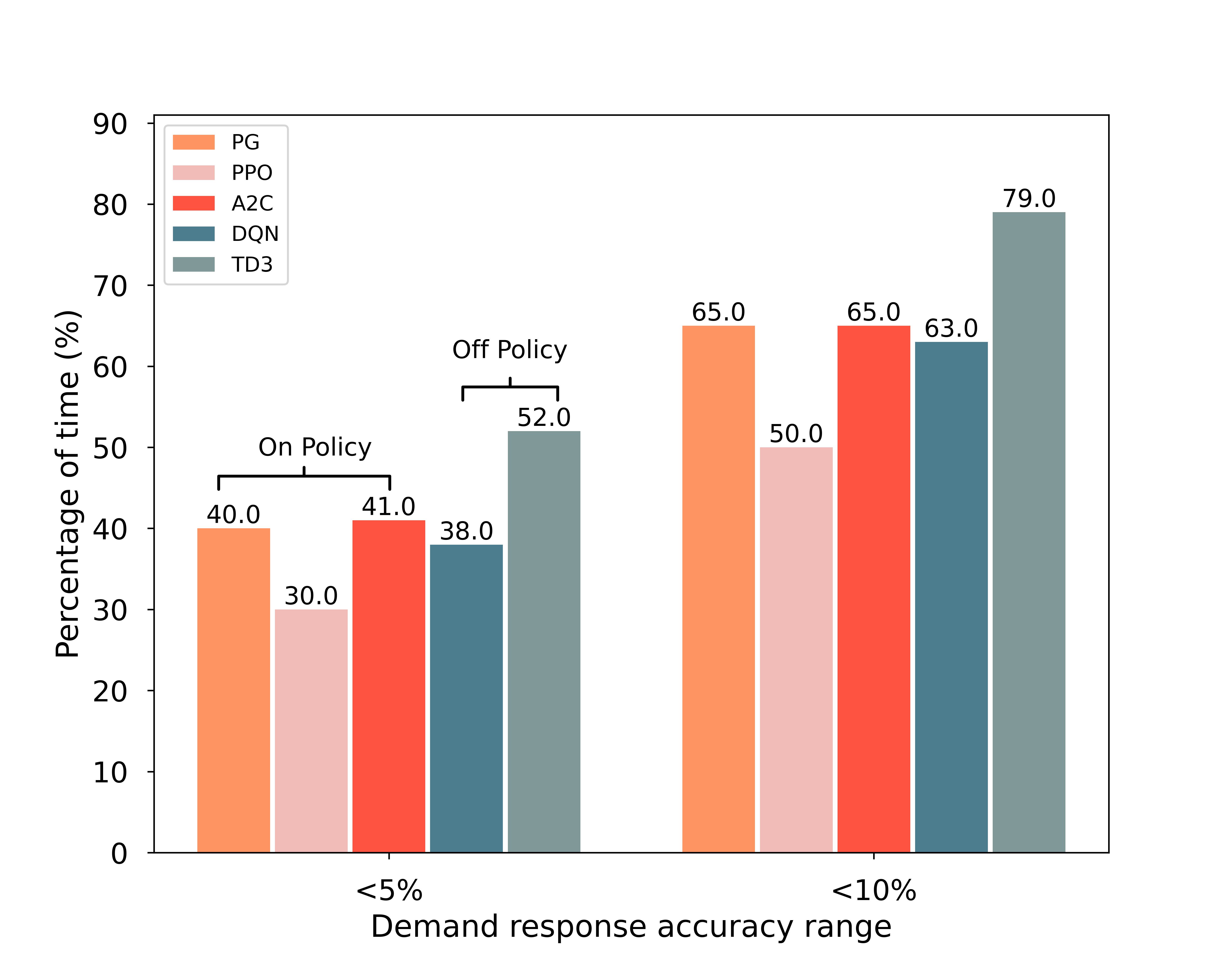}
    \caption{Percentage of time in different error groups for constant demand response control}
    \label{constant_barplot}
\end{figure}

To gain better insight into the accuracy of the responses, we present the percentage of time where the accuracy error is within 5\% and 10\% for all algorithms. As shown in Figure \ref{constant_barplot}, off-policy algorithms outperform on-policy algorithms. Among them, TD3 demonstrates the best performance under these criteria, with 53\% of time having response errors below 5\% and 82\% below 10\%. In contrast, the three on-policy algorithms exhibit similar performance, with approximately 35\% of time showing control errors less than 5\% and around 60\% less than 10\%. TD3 stands out with the best performance in maximizing time with control error less than 10\%.

\begin{figure}
    \centering
    \includegraphics[width=1\linewidth]{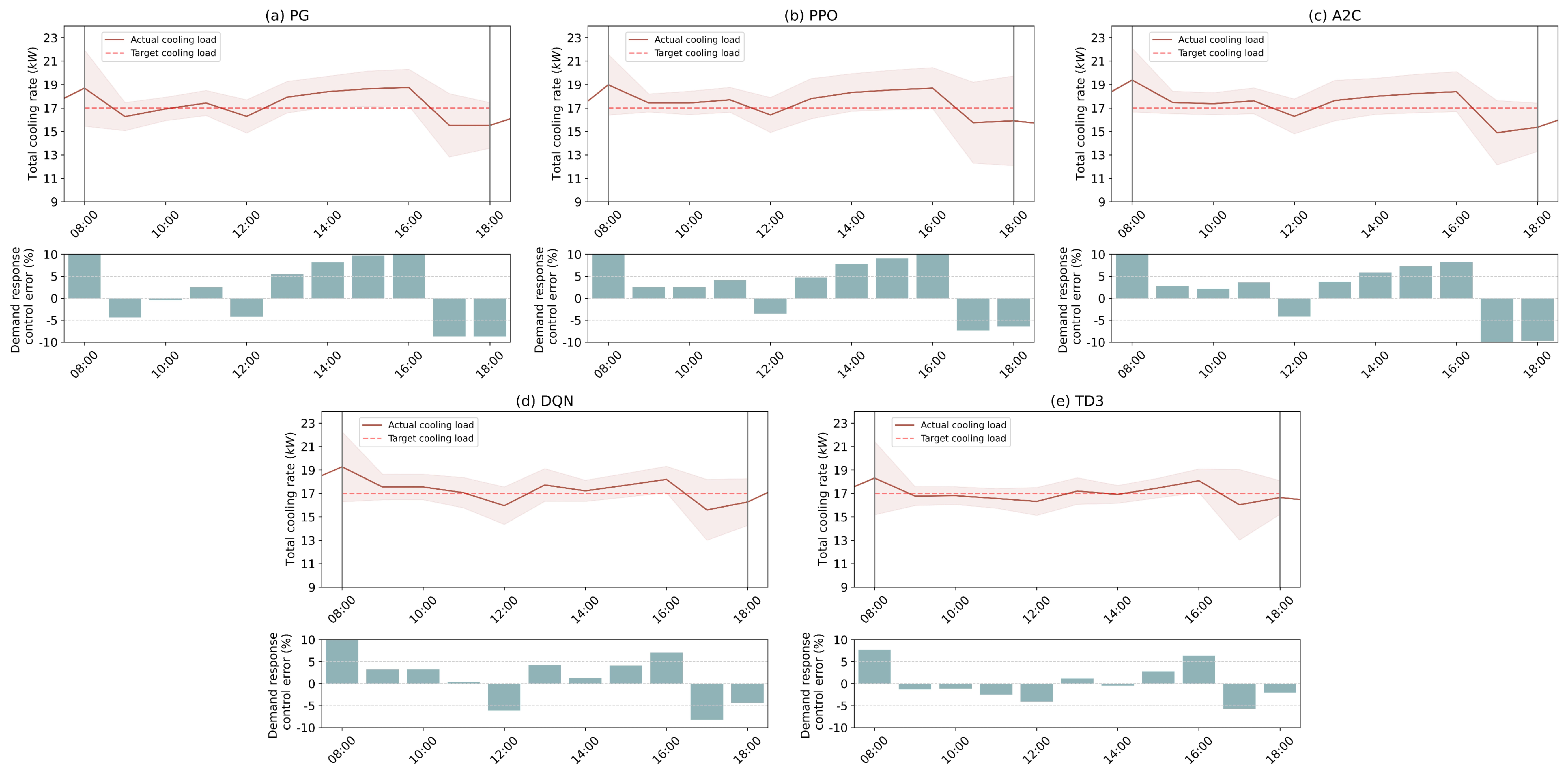}
    \caption{The daily averaged cooling load profile with one stand interval and the mean control error at each time of days under constant demand reduction (a: Policy gradient, b: Proximal policy optimization, c: Advantage actor critic, d: Double deep Q-learning,  d: TD3)}
    \label{constant_cooling_energy}
\end{figure}

Figure \ref{constant_cooling_energy} illustrates the cooling energy consumption profiles along with the corresponding standard deviation for each RL algorithm. Under RL control, the cooling energy profiles are significantly flattened due to adjustments in the indoor cooling setpoint over time. Notably, the standard deviation range fully covers the target cooling energy at each time of the day, indicating the effectiveness of the RL control strategy throughout the day. In comparison to the baseline (Figure \ref{fig:baseline}), the uncertainty in cooling energy consumption is reduced by 40\% to 50\% under the RL control strategy. This further demonstrates that RL can effectively regulate cooling energy within the desired range by adjusting indoor cooling setpoints while considering real-time indoor and outdoor environmental conditions.

From the view of hourly averaged cooling load management for the whole simulation period, TD3 and DQN outperform the other RL algorithms, maintaining the hourly average control errors below 5\% during most periods. However, the demand control error is slightly higher at 8:00 AM, 12:00 PM, and after 5:00 PM, primarily due to significant changes in occupancy and lighting/equipment schedules at these times. The RL algorithms struggle to respond quickly because they rely solely on real-time information rather than predicted data. In regard to on-policy algorithms, they exhibits better performance in the afternoon compared to the morning. 

\begin{figure}[th]
    \centering
    \includegraphics[width=1\linewidth]{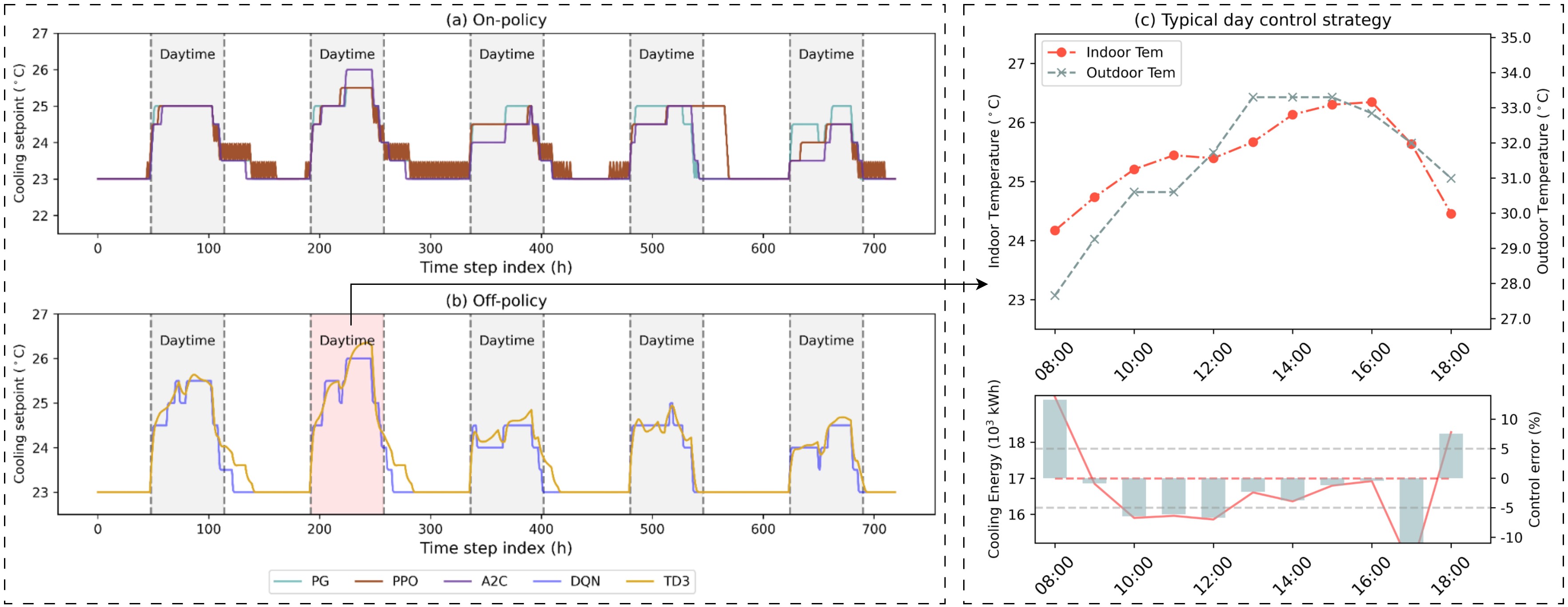}
    \caption{The indoor temperature profile under different RL strategies for constant cooling reduction management (typical day control schedule is under TD3 strategy)}
    \label{constant_typical}
\end{figure}

\subsubsection{Typical day control}


Figure \ref{constant_typical} (a) and (b) present the control schedules for constant reduction under on-policy and off-policy algorithms across five consecutive working days, respectively. It is interesting to note that the control schedule trends for off-policy algorithms are similar, despite the fact that DQN operates in a discrete action space while TD3 operates in a continuous action space. The performance difference between TD3 and DQN can likely be attributed to their differing action spaces, as the continuous action space enables more precise control. In contrast, the control strategies for on-policy algorithms exhibit obvisous differences on certain days. Given the lower performance of on-policy algorithms, this inconsistency suggest that they may struggle to converge to a globally optimal strategy, even though they reduce uncertainty of cooling load to some extent.

Figure \ref{constant_typical}(c) presents the control schedule for a typical day. In the morning, the control algorithm gradually increases the cooling setpoint from 23 $^\circ$C to 26 $^\circ$C as the outdoor temperature rises, continuing this trend until 3:00 PM. Afterward, the cooling setpoint begins to decrease slightly to 24 $^\circ$C. Throughout this day, the cooling energy is successfully maintained at the desired load of 17 kW, with an error of less than 5\% for most of the time. Based on this control schedule and the corresponding cooling load profile, it is evident that the RL algorithm can learn the thermal dynamics of buildings to regulate cooling load effectively for long-term demand response control. Additionally, the indoor temperature range is approximately 23.5 $^\circ$C to 25.5 $^\circ$C, indicating that the cooling load can be maintained at the desired level  with acceptable indoor thermal comfort. Notably, the peak indoor temperature lags about 2 hours behind the outdoor temperature. This suggests that the RL strategy successfully learns the characteristics of building thermal dynamics to regulate the cooling load within the desired range rather than relying solely on outdoor temperature.

\subsection{Cooling load management with dynamic target}\label{dynamic_demand}

The problem of cooling load management with dynamic target is an extension of cooling load management of constant target. The difference between dynamic target and the constant target is that an external signal is introduced in the dynamic target problem to simulate the changing demand response requirement from grid, while the object of constant target is to flatten the cooling load profile and no external signal is considered. This enhancement enables more flexible cooling load management, allowing buildings to function as dynamic demand regulators. By showcasing this control problem configuration, we aims to validate the ability of BuildingGym in considering external control signals for application in grid-interactive buildings.

\subsubsection{Basic configuration}

Since the control problem is the extension of previous problem, the configuration is basically same. The key difference is that we define the external signal function to simulate the change of demand response requirement from grid.

In this control problem, an external signal indicates the desired cooling load level on an hourly basis. The response schedule is outlined in Table \ref{signal}. The signal values vary within the set \(\{0, 0.5, 1\}\), which indicates the cooling load reduction of $\{0\%, 15\%, 30\%\}$, respectively. To enable the RL algorithm to understand the desired cooling reduction level for each hour, we have included this response signal as an additional observation input. The other inputs and action functions remain unchanged from those used in the constant reduction problem.

\begin{table}[htb]
\caption{The schedule of external signal for desired reduction in cooling load}\label{signal}
\begin{tabular}{p{3cm}p{3cm}p{3cm}}
\toprule
Time          & External signal & Reduction target \\
\midrule
8:00 - 11:00  & 0               & 0\%              \\
11:00 - 13:00 & 1               & 30\%             \\
13:00 - 14:00 & 0               & 0\%              \\
14:00 - 16:00 & 0.5             & 15\%             \\
16:00 - 17:00 & 0               & 0\%              \\
17:00 - 19:00 & 1               & 30\%            \\
\bottomrule
\end{tabular}
\end{table}

\subsubsection{Control accuracy under dynamic externals}

\begin{figure}[t]
    \centering
    \includegraphics[width=0.8\linewidth]{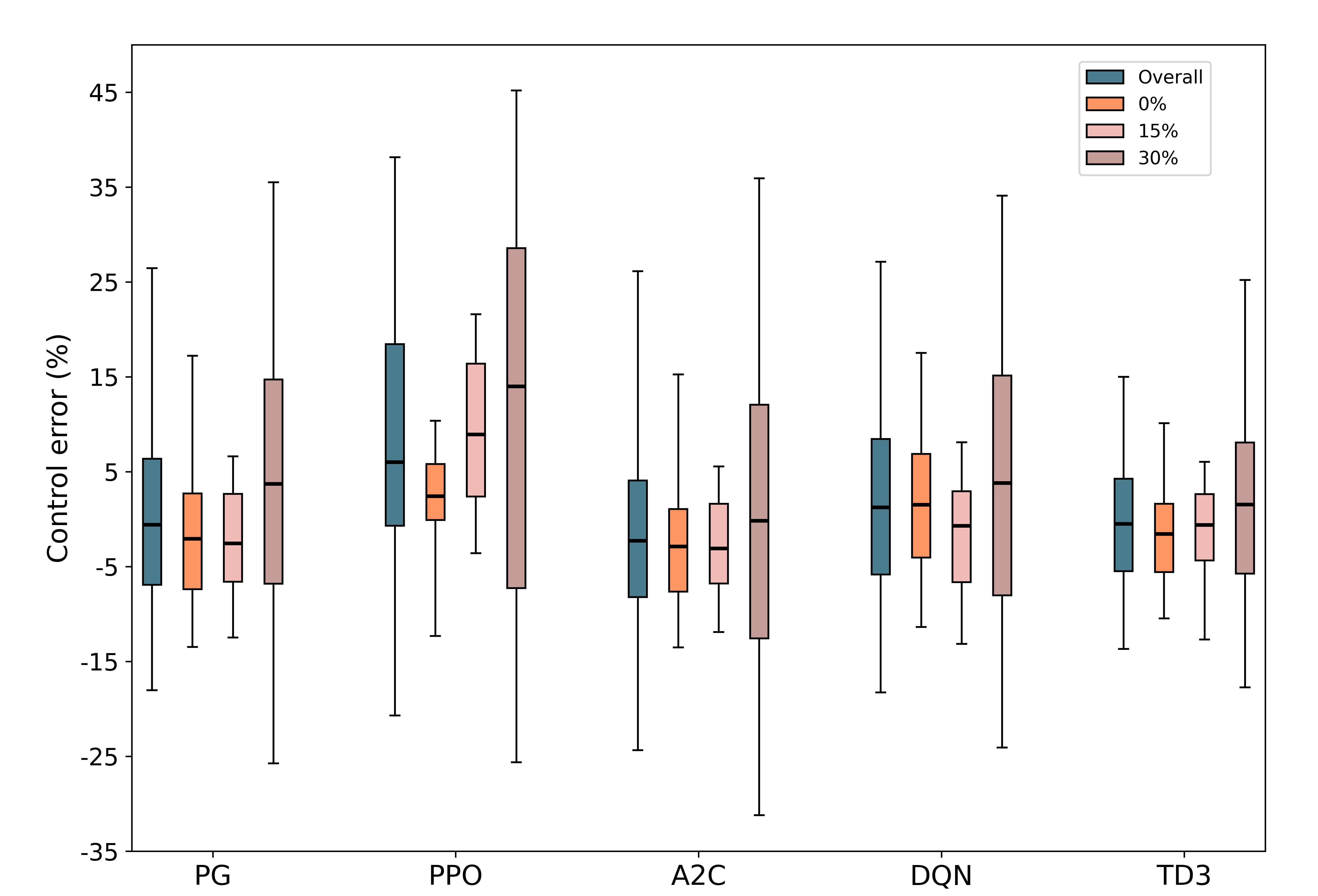}
    \caption{The box plot of control error in different reduction signals for each RL}
    \label{variable_boxplot}
\end{figure}

In this section, we present the performance of trained RL agent for cooling load management with dynamic target to validate the effectiveness of BuildingGym in considering external control signals from communities.

Figure \ref{variable_boxplot} shows the box plots of overall control error, as well as the error within groups of reduction signals for each RL algorithm. In terms of overall control error, four of the RL algorithms maintain an average control error within \(\pm 2\%\), while only PPO exhibits a relatively larger average control error of 6\%. All RL algorithms successfully maintain median control error within \(\pm 3\%\) during periods with a response signal of 0. For the time with a control signal of 1, the median control errors of the two off-policy algorithms, TD3 and DQN, are -1\%, while PG and A2C have a median control error of -3\%.  However, it is interesting to note that although the A2C algorithms achieves a median control error of 0\% during the time with a response signal of 2, the corresponding range between 10th and 90th is also the largest. This observation indicates that for this control task, A2C perform well in long-term evaluation but not in real-time evaluation.

\begin{figure}[t]
    \centering
    \includegraphics[width=0.7\linewidth]{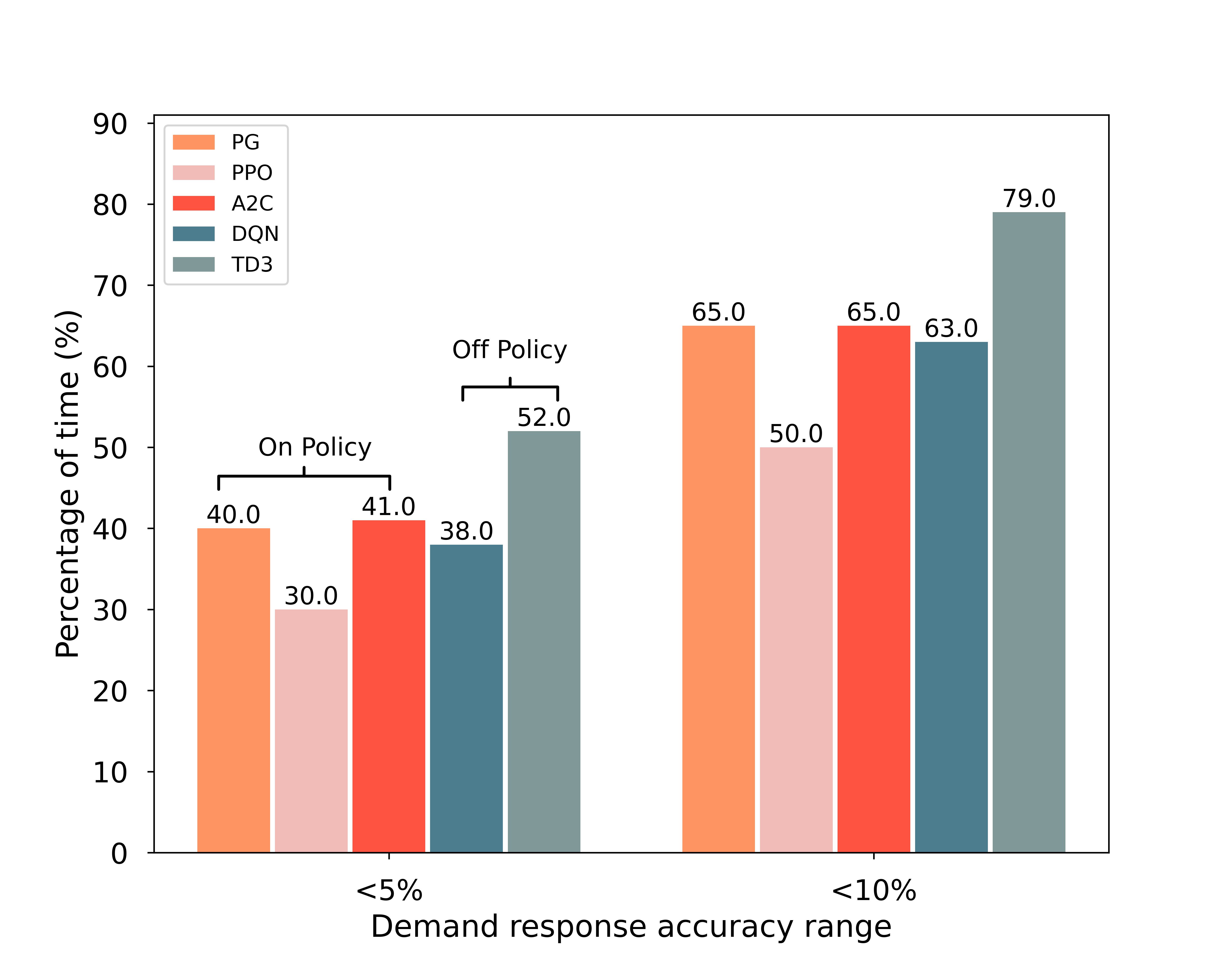}
    \caption{Percentage of time in different error groups for dynamic demand response control}
    \label{variable_barplot}
\end{figure}

Figure \ref{variable_barplot} shows the percentage of time across different control error groups for the dynamic demand reduction problem. Among the on-policy algorithms, PG and A2C perform better than PPO. There is about 65\% of time with control error less than 10\% for these two algorithms while it is 50\% under the PPO algorithm. In comparing on-policy and off-policy algorithms, DQN shows similar performance to PG and A2C. However, TD3 outperforms all algorithms, achieving 52\% of time with a control error of less than 5\% and 79\% with an error of less than 10\%. It is interesting to note that DQN, as an off-policy learning algorithm, does not significantly outperform the on-policy algorithms in this dynamic target problem. This indicates that off-policy methods are not guaranteed to provide better control performance than on-policy methods.

\begin{figure}[htb]
    \centering
    \includegraphics[width=1\linewidth]{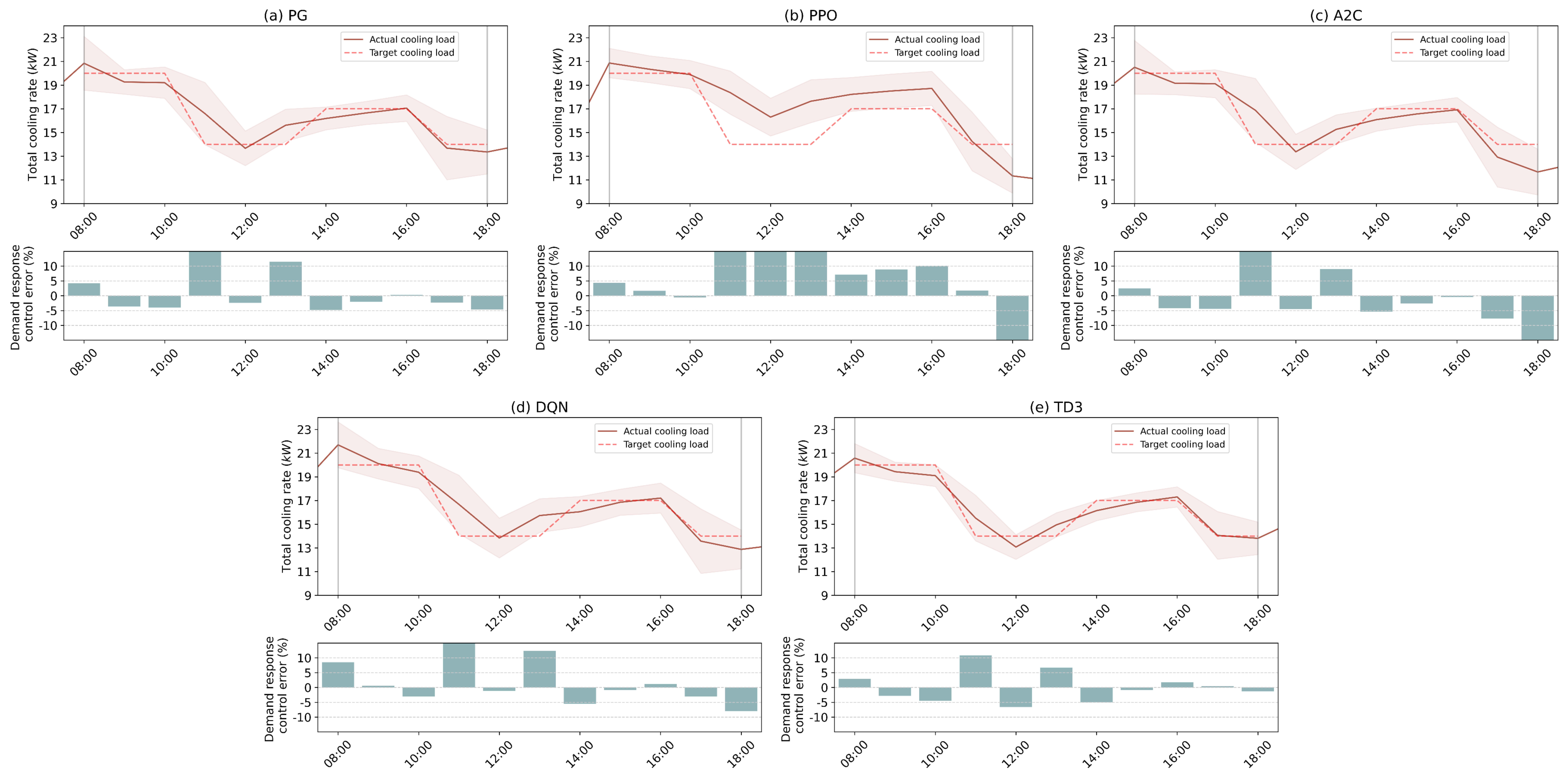}
    \caption{The daily averaged cooling load profile with one stand interval and the mean control error at each time of days under dynamic demand reduction (a: Policy gradient, b: Proximal policy optimization, c: Advantage actor critic, d: Double deep Q-learning,  d: TD3)}
    \label{cooling_load_var}
\end{figure}

Figure \ref{cooling_load_var} shows the hourly average cooling load profile under each algorithm. The solid lines indicate the average cooling rate in each hour and the dash lines indicate the target cooling rate under the schedule of reduction signals. Except for PPO, the cooling load profiles under the other four algorithms show similar trends required by external signals, showing a good performance of RL in regulating cooling load for demand response control. In general, the cooling load slightly decreased from 19 kW to 13 kW in the morning. Starting at 12:00 PM, the cooling loads exhibit a slight rebound trend with the cooling rate increasing to about 17 kW until 4:00 PM. After that, the loads change to decrease again as the desired reduction rate becomes 30\% after 5:00 PM. In terms of hourly average cooling rate management, the off-policy algorithms (TD3 and DQN) show better performance than on-policy learning algorithms. 

\subsubsection{Typical day control}

\begin{figure}[htbp]
    \centering
    \includegraphics[width=1\linewidth]{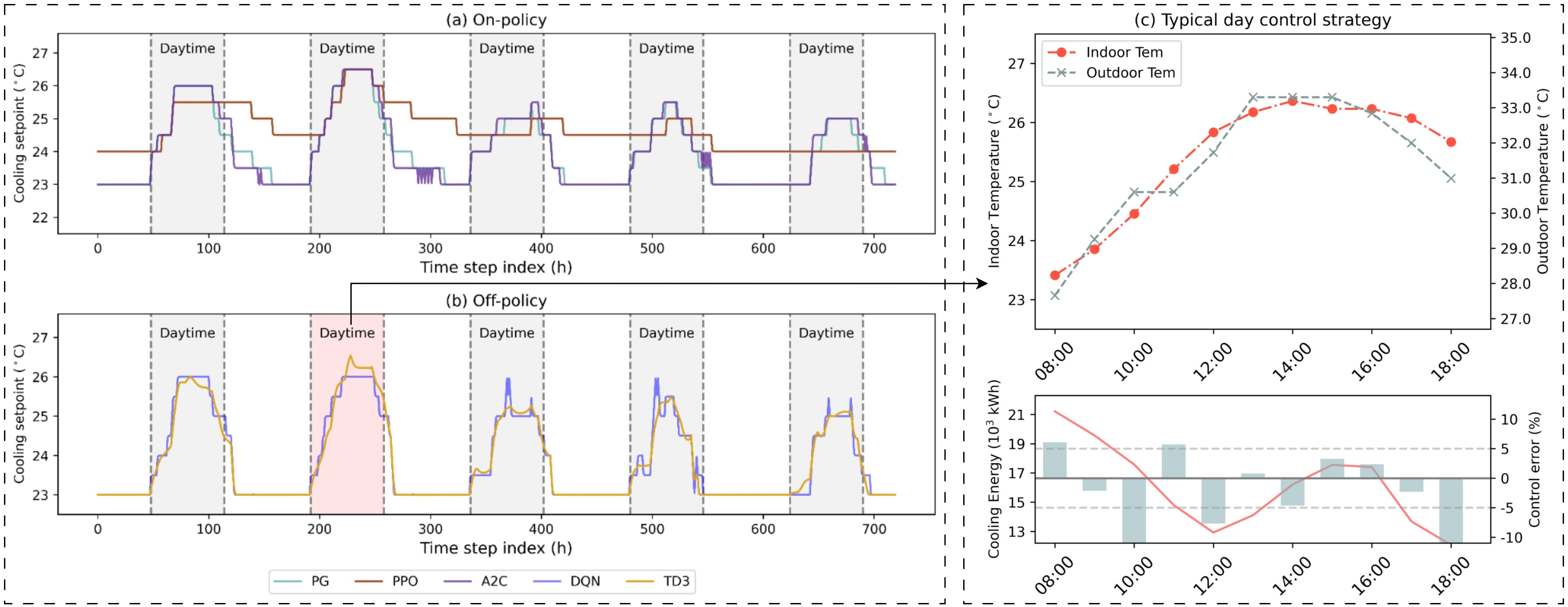}
    \caption{The indoor temperature profile under different RL strategies for dynamic cooling reduction management (Typical day control schedule is under TD3 strategy)}
    \label{variable_typical}
\end{figure}

Figure \ref{variable_typical} (a) and (b) show the control schedules for five consecutive working days using on-policy and off-policy algorithms, respectively. Compared to constant reduction (Figure \ref{constant_typical}), the indoor temperature range during the daytime is wider due to the larger cooling load reduction required by the dynamic target. Under the control with dynamic target, the control schedules of PG, A2C, and DQN are similar, resulting in comparable performance as shown in Figure \ref{variable_barplot}. The superior performance of TD3 may be attributed to its more precise control enabled by continuous action spaces. The PPO algorithm appears to struggle in training an agent that matches the performance of the other four algorithms.

Figure \ref{variable_typical}(c) illustrates the control schedule for following the real-time external response signal on cooling rate under the TD3 strategy. Compared to the schedule with a constant cooling target (shown in Figure \ref{constant_typical}), the indoor temperature peaks earlier at around 1:00 PM, due to a greater cooling load reduction requirement at noon. This observation indicates that the RL algorithm effectively combines external control signals with building thermal dynamics to regulate cooling load at the desired level. For this typical day, the indoor temperature ranges from 23 $^\circ$C to 27 $^\circ$C, and the control error remains within 10\% for over 80\% time of the day. This further suggests that it is possible to manage the indoor cooling load within a reasonable range while maintaining a satisfactory indoor environment.

\section{Discussion}

In this paper, we propose BuildingGym for AI-based building energy management. With BuildingGym, building managers can efficiently set up the training task for a wide range of control problem related to building energy system, while AI specialists can test the any RL algorithms using the demo case provided by BuildingGym. Therefore, a key feature of this toolbox is its ability to bridge the gap between AI specialists and building managers, facilitating better collaboration between these two groups in the building energy field. In this section, we discuss the strength of BuildingGym and the potential improvement in future releases. 

\subsection{Bridging the gap between AI specialists and building managers}

In the current release, we integrate EnergyPlus as the simulator for building simulations, providing comprehensive building observation data and control variables for common energy systems. We have also established an easy-to-use API between EnergyPlus and Python, allowing users to easily select from a list of observation and control variables. One of the contributions of BuildingGym is that it helps to bridge the gap between AI specialists and building managers. For AI specialists, they can focus on optimizing RL algorithms in BuildingGym and test the proposed RL algorithm using built-in training environments provided by BuildingGym. If they get an algorithm that is fully developed and ready for testing, they simply need to choose the relevant inputs (observations) and outputs (control variables) from the list to evaluate the control performance of their newly developed AI algorithms. In the other hands, for building managers, we offer five built-in RL algorithms. These algorithms have been fine-tuned and tested in a virtual testbed building, demonstrating strong performance in various applications. They can easily replace the built-in virtual testbed with their own building models and define the RL training environment according to their specific requirements and get the optimized RL agent for the specific task using BuildingGym.
Therefore, both AI specialists and building managers can focus on their specialty and contribute to optimization control for building energy systems through BuildingGym.


Additionally, since EnergyPlus provides comprehensive building simulation data and can simulate various types of energy systems, building managers can easily define their control problems/target through reward functions to develop the corresponding AI control agents using the built-in algorithms. Also, it should be noted that the simulator of the BuildingGym is replaceable. Building managers are able to choose their own software, e.g. DeST \citep{yan2022dest}, as long as the simulators can provide observation and control API at each time step. 

It should be noted that the simulation engine in BuildingGym is replaceable to other simulation engines or connecting to actual building BMS system, providing more possibilities for implementation in real buildings. In future release of BuildingGym, it is interesting to adapt BuildingGym API to real building BMS though OSIsoft, with which the building operation information get be easily retrieved, and the action can be sent to BMS in Python interface. By doing this, we can further test the performance of RL in real building control.

\subsection{Designed for buildings under energy transition}

BuildingGym is designed to be flexible with simulators. Any simulator capable of providing observations and receiving control commands in real-time can be easily integrated into BuildingGym. This flexibility allows BuildingGym to work as a control optimization toolbox for real buildings by collaborating with building energy systems. Moreover, BuildingGym is able to accept the external variable as the observation for RL training. This feature makes BuildingGym convenient applicable for grid-response buildings in complex energy system, which would be an important tool for building energy management in energy transition.

\subsection{Selecting suitable RL algorithm}

\citet{wang2020reinforcement} reviews studies on RL for building energy system control and finds that policy gradient methods and deep Q-networks (DQN) are the most frequently used algorithms in this field, primarily due to their ease of implementation. However, our example application demonstrates that TD3 outperforms the others in cooling load management. While case studies may not provide definitive comparisons of algorithm performance, it is still reasonable to say testing several RL algorithms and selecting the best one for the specific control problem is important. To support this, we offer several built-in RL algorithms for users to choose from and evaluate. Moving forward, we plan to continuously update and integrate state-of-the-art RL algorithms in future releases.

\subsection{Multi-agent control and automatic fine-tune}

For the entire building energy system, controlling both room-level and system-level setpoints can significantly enhance overall building efficiency. This approach allows the system to be optimized to achieve the highest efficiency for terminal rooms according to their actual working conditions. The multi-agent RL strategy is designed to simultaneously control multiple agents and facilitate cooperation among them. While we present single-agent control for cooling load management in these examples, it is important to note that BuildingGym also supports decentralized multi-agent control problems.

In the current release of BuildingGym, we offer a batch file template for fine-tuning RL algorithms. The agent demonstrating the best control performance is automatically saved during training. Additionally, all training processes can be visualized online, providing developers with better insights into the training progression. Currently, several automatic fine-tuning technologies exist to accelerate the fine-tuning process and help identify optimal hyperparameters for specific control problems. In future releases, we plan to integrate these automatic fine-tuning technologies into BuildingGym to further enhance training efficiency.


As the share of renewable energy increases in the energy market, buildings must operate more flexibly to respond to grid requirements. Several technologies are being integrated into buildings to enhance this flexibility. For instance, batteries can operate in charge/discharge mode to adjust load profiles, and with the growing number of electric vehicles, the batteries in these vehicles can serve as valuable distributed energy storage sources for buildings. However, the operation of such distributed energy storage systems presents several challenges. One significant challenge is the stochastic nature of vehicle parking times, meaning these batteries are not always available for building use. Additionally, the state of charge in EVs must be sufficient to meet travel demands when the vehicles are driven away. Effective negotiation and coordination between building managers and electric vehicle owners are also crucial. These challenges highlight the urgent need for a standardized environment that can facilitate the integration of these multi-source energy communities, which we plan to implement in future releases of BuildingGym.

\section{Conclusions}

In this study, we developed BuildingGym, an open-source tool for AI-based building energy management. By integrating EnergyPlus with RL, BuildingGym bridges the gap between building managers and AI specialists. Additionally, BuildingGym can accept external signals as part of its input, enabling more flexible control that aligns with grid demand response requirements.

We tested BuildingGym’s effectiveness in managing cooling loads with both constant and dynamic targets. With BuildingGym, we efficiently set up training tasks for cooling load management. The training results demonstrate that the control strategies trained by BuildingGym effectively account for building physics dynamics and external requirements, successfully regulating the cooling load within the desired range. This highlights the effectiveness of BuildingGym in AI-based building energy management.

\section*{Acknowledgements}

This research is supported by AI Singapore under its project of Development of NetZero BEMS through AI-based HVAC System Control (AISG2-TC-2023-008-SGKR). 

\begingroup

\bibliography{sn-bibliography}

\end{document}